\documentclass[11pt]{article}

\usepackage{times}
\usepackage{graphicx} 
\usepackage{subfigure}

\usepackage{algorithm}
\usepackage{algorithmic}

\usepackage{amsmath}
\usepackage{amsfonts}
\usepackage{amssymb}

\newcommand{\scU}{\mathcal{U}}

\newcommand{\hj}{\ensuremath{\widehat{j}}}
\newcommand{\bbw}{\ensuremath{\bar{\bw}}}
\newcommand{\bM}{\ensuremath{\bar{M}}}
\newcommand{\bbb}{\ensuremath{\bar{\bb}}}

\newcommand{\field}[1]{\mathbb{#1}}

\newcommand{\bb}{\boldsymbol{b}}

\newcommand{\bx}{\boldsymbol{x}}

\newcommand{\bw}{\boldsymbol{w}}
\newcommand{\bu}{\boldsymbol{u}}

\newcommand{\bzero}{\boldsymbol{0}}

\newcommand{\R}{\field{R}}

\newcommand{\scO}{\mathcal{O}}

\renewcommand{\Pr}{\mathbb{P}}

\newcommand{\ignore}[1]{}

\DeclareMathOperator*{\argmax}{argmax}

\newcommand{\CB}{\mbox{\sc cb}}
\newcommand{\sCB}{\widetilde \CB}

\begin{document}

\title{\bf Graph Clustering Bandits for Recommendation}

\author {Shuai Li\\DiSTA, University of Insubria, Italy\\
\texttt{shuaili.sli@gmail.com}
\and
Claudio Gentile\\DiSTA, University of Insubria, Italy\\
\texttt{claudio.gentile@uninsubria.it}
\and
Alexandros Karatzoglou\\
Telefonica Research, Spain\\
\texttt{alexk@tid.es}
}

\maketitle
\begin{abstract}
We investigate an efficient context-dependent clustering technique for recommender systems based on exploration-exploitation strategies through multi-armed bandits over multiple users. Our algorithm dynamically groups users based on their observed behavioral similarity during a sequence of logged activities. In doing so, the algorithm reacts to the currently served user by shaping clusters around him/her but, at the same time, it explores the generation of clusters over users which are not currently engaged. We motivate the effectiveness of this clustering policy, and provide an extensive empirical analysis on real-world datasets, showing scalability and improved prediction performance over state-of-the-art methods for sequential clustering of users in multi-armed bandit scenarios.
\end{abstract}

\section{Introduction}
Exploration-exploitation techniques a.k.a. Bandits are becoming an essential tool in modern recommenders systems \cite{recsys13,recsys14}. Most recommendation setting involve an ever changing dynamic set of items, in many domains such as news and ads recommendation the item set is changing so rapidly that is impossible to use standard collaborative filtering techniques. In these settings bandit algorithms such as contextual bandits have been proven to work well \cite{lcls10} since they provide a principled way to gauge the appeal of the new items.

Yet, one drawback of contextual bandits is that they mainly work in a content-dependent regime, the user and item content features determine the preference scores so that any collaborative effects (joint user preferences over groups of items) that arise are being ignored. Incorporating collaborative effects into bandit algorithms can lead to a dramatic increase in the quality of recommendations. In bandit algorithms this has been mainly done by clustering the user. 
For instance, we may want to serve content to a group of users by taking advantage of
an underlying network of preference relationships among them. These preference relationships can either be
explicitly encoded in a graph, where adjacent nodes/users are deemed similar to one another, or implicitly
contained in the data, and given as the outcome of an inference process that recognizes similarities
across users based on their past behavior.

\sloppypar{
To deal with this issue a new type of bandit algorithms has been developed which work under the assumption that users can be grouped (or clustered) based on their selection of items e.g. \cite{icml14,cikm14}. The main assumption is that users form a graph where the edges are constructed based on signals of similar preference (e.g. a number of similar item selections). By partitioning the graphs, one can find a coherent group of users with similar preference and online behavior. While these algorithms have been proven to perform significantly better then classical contextual bandits, they "only" provide exploration on the set of items. For new users or users with sparse activity it is very difficult to find an accurate group assignment. This is a particularly important issue since most recommendation settings face unbalancedness of user activity levels, that is, relatively few users have very high activity while the vast majority of users have little to practically no activity at all (see Figure \ref{fig:activity_level} in Section \ref{s:exp}).
}

In this work, we introduce a new bandit algorithm that adds an extra exploration component over the group of users. In addition to the standard exploration-exploitation strategy over items, this algorithm explores different clustering assignments of new users and users with low activity. The experimental evaluation on four real datasets against baselines and state-of-the-art methods confirm that the additional dynamic paradigm tends to translate into solid performance benefits.

\section{Learning Setting}\label{s:setting}
As in previous work in this literature (e.g., \cite{icml14,cgz13}), we assume that user behavior similarity is represented by an underlying (and unknown) clustering over the set of users.
Specifically, if we let $\scU = \{1,\ldots, n\}$ be the set of $n$ users, we assume $\scU$ can be partitioned
into a small number $m$ of clusters $U_1, U_2, \ldots, U_{m}$, where $m$ is expected to be much smaller than $n$.
The meaning of this clustering is that users belonging to the same cluster $U_j$ tend to have similar behavior, while
users lying in different clusters have significantly diverging behavior.
Both the partition $\{U_1, U_2, \ldots, U_{m}\}$ and the common user behavior within each cluster are {\em unknown}
to the learning system, and have to be inferred on the fly based on past user activity.

The above inference procedure has to be carried out within a sequential decision setting where the learning system (henceforth ``the learne'') has to continuously adapt to the newly received information provided by the users. To this effect, the learning process is divided into a discrete sequence of rounds:  in round $t=1,2,\dots$, the learner receives a user index $i_t \in \scU$ to serve content to. Notice that
the user to serve may change at every round, though the same user can recur many times. The sequence $\{i_t\}$ is exogenously determined by the way users interact with the system, and is not under our control. In practice, very high unbalancedness levels of user activity are often observed. Some users are very active, many others are (almost) idle or newly registered users, and their preferences are either extremely sporadic or even do not exist~\cite{recsys13}. Along with $i_t$, the system receives in round $t$ a set of feature vectors $C_{i_t} = \{\bx_{t,1}, \bx_{t,2},\ldots, \bx_{t,c_t}\} \subseteq \R^d$ representing the content which is currently available for recommendation to user $i_t$. The learner picks some ${\bar \bx_t} = \bx_{t,k_t} \in C_{i_t}$ to recommend to $i_t$, and then observes $i_t$'s feedback in the form of a numerical payoff $a_t \in \R$. In this scenario, the learner's goal is to maximize its total payoff\, $\sum_{t=1}^T a_t$\, over a given number $T$ of rounds. 
When the user feedback the learner observes is only the click/no-click behavior, the payoff $a_t$ can be naturally interpreted
as a binary feedback, so that the quantity
\[
\frac{\sum_{t=1}^T a_t}{T}
\]
becomes a click-through rate (CTR), where
$a_t = 1$ if the recommended item ${\bar \bx_t}$ was clicked by user $i_t$, and $a_t = 0$, otherwise. In our experiments (Section \ref{s:exp}), when the data at our disposal only provide the payoff associated with the item recommended
by the logged policy,
CTR is our measure of prediction accuracy. On the other hand, when the data come with payoffs for all possible items in $C_{i_t}$, then our measure of choice will be the cumulative {\em regret} of the learner,\footnote
{
In fact, for the sake of clarity, our plots will actually display {\em ratios} of cumulative regrets and {\em ratios} of CTRs ---see Section \ref{s:exp} for details.
}
defined as follows. Let $a_{t,k}$ be the payoff associated in the data at hand with item $\bx_{t,k} \in C_{i_t}$. Then the regret $r_t$ of the learner at time $t$ is the extent to which the payoff of the best choice in hindsight at user $i_t$ exceeds the payoff of the learner's choice, i.e.,
\[
r_t = \Bigl(\max_{k = 1, \ldots, c_t}\, a_{t,k} \Bigl) - a_{t,k_t}~,
\]
and the cumulative regret is simply
\[
\sum_{t=1}^T r_t~.
\]
Notice that the round-$t$ regret $r_t$ refers to the behavior of the learner when predicting preferences of user $i_t$, thus the cumulative regret takes into duly account the relative ``importance" of users, as quantified by their activity level. The same claim also holds when measuring performance through CTR.

\section{The Algorithm}\label{s:alg}
%
Our algorithm, called Graph Cluster of Bandits (GCLUB, see Figure \ref{alg:gclub}), is a variant of the Cluster of Bandits (CLUB) algorithm originally introduced in \cite{icml14}. We first recall how CLUB works, point out its weakness, and then describe the changes that lead to the GCLUB algorithm.

The CLUB algorithm maintains over time a partition of the set of users $\scU$ in the form of {\em connected components} of an undirected graph $G_t =(\scU,E_t)$ whose nodes are the users, and whose edges $E_t$ encode our current belief about user similarity. CLUB starts off from a randomly sparsified version of the complete graph (having $\scO(n\log n)$ edges instead of $\scO(n^2)$), and then progressively deletes edges based on the feedback provided by the current user $i_t$. Specifically, each node $i$ of this graph hosts a linear function $\bw_{i,t}\,: \bx \rightarrow \bw_{i,t}^\top\bx$ which is meant to estimate the payoff user $i$ would provide had item $\bx$ been recommended to him/her. The hope is that this estimates gets better and better over time. When the current user is $i_t$, it is only vector $\bw_{i_t,t}$ that is updated based on $i_t$'s payoff signal, similar to a standard linear bandit algorithm (e.g.,~\cite{Aue02,chu2011contextual,abbasi2011improved,cgz13}) operating on the context vectors contained in $C_{i_t}$. Every user $i \in \scU$ hosts such a linear bandit algorithm. The actual recommendation issued to $i_t$ within $C_{i_t}$ is computed as follows. First, the connected component (or cluster) that $i_t$ belongs to is singled out (this is denoted by $\hj_t$ in Figure \ref{alg:gclub}). Then, a suitable aggregate prediction vector (denoted by $\bbw_{\hj_t,t-1}$ in Figure \ref{alg:gclub}) is constructed which collects information from all uses in that connected component. The vector so computed is engaged in a standard upper confidence-based exploration-exploitation tradeff to select the item ${\bar \bx_t} \in C_{i_t}$ to recommend to user $i_t$. 

Once $i_t$'s payoff $a_t$ is received, $\bw_{i_t,t-1}$ gets updated to $\bw_{i_t,t}$ (through $a_t$ and ${\bar \bx_t}$). In turn, this may change the current cluster structure, for if $i_t$ was formerly connected to, say, node $\ell$ and, as a consequence of the update $\bw_{i_t,t-1} \rightarrow \bw_{i_t,t}$, vector $\bw_{i_t,t}$ is no longer close to $\bw_{\ell,t-1}$, then this is taken as a good indication that $i_t$ and $\ell$ cannot belong to the same cluster, so that edge $(i_t,\ell)$ gets deleted, and new clusters over users are possibly obtained.

The main weakness of CLUB in shaping clusters is that when responding to the current user feedback, the algorithm operates only {\em locally} (i.e., in the neigborhood of $i_t$). While advantageous from a computational standpoint, in the long run this has the severe drawback of overfocusing on the (typically few) very active users; the algorithm is not responsive enough to those users on which not enough information has been gathered, either because they are not so active (typically, the majority of them) or because they are simply newly registered users. In other words, in order to make better recommendations it's worthy to discover and capture the ``niches'' of user preferences as well. 

Since uneven activity levels among users is a standard pattern when users are many (and this is the case with our data, too --- see Section \ref{s:exp}), GCLUB complements CLUB with a kind of stochastic exploration at the level of cluster shaping. In every round $t$, GCLUB deletes edges in one of two ways: with independent high probability $1-r > 1/2$, GCLUB operates on component $\hj_t$ as in CLUB, while with low probability $r$ the algorithm picks a connected component uniformly at random among the available ones at time $t$, and splits this component into {\em two} subcomponents by invoking a fast graph clustering algorithm (thereby generating  one more cluster). This stochastic choice is only made during an initial stage of learning (when $t \leq T/10$ in GCLUB's pseudocode), which we may think of as a {\em cold start} regime for most of the users.
The graph clustering algorithm we used in our experiments was implemented through the {\em Graclus} software package from the authors of \cite{pami07,sigkdd05}. Because we are running it over a single connected component of an initially sparsified graph, this tool turned out to be quite fast in our experimentation. 

The rationale behind GCLUB is to add an extra layer of exploration-vs-exploitation tradeoff, which operates at the level of clusters.
At this level, exploration corresponds to picking a cluster at random among the available ones, while exploitation corresponds to working on the cluster the current user belongs to. In the absence of enough information about the current user and his/her neighborhood, exploring other clusters is intuitively beneficial. We will see in the next section that this is indeed the case.
\begin{figure}[t!]
\begin{center}
\begin{algorithmic}
\small
\STATE \textbf{Input}: Exploration parameter $\alpha > 0$; 
cluster exploration probability $r < 1/2$.
\STATE \textbf{Init}:
\vspace{-0.1in}
\begin{itemize}
\item $\bb_{i,0} = \bzero \in \R^d$ and $M_{i,0} = I \in \R^{d\times d}$,\ \ $i = 1, \ldots n$;
\vspace{-0.1in}
\item Clusters ${\hat U_{1,1}} = \scU$, number of clusters $m_1 = 1$;
\vspace{-0.1in}
\item Graph $G_1 = (\scU,E_1)$, $G_1$ has $\scO(n\log n)$ edges and is connected.
\end{itemize}
\vspace{-0.1in}
\FOR{$t =1,2,\dots,T$}
\STATE Receive $i_t \in \scU$;
\STATE Set $\bw_{i,t-1} = M_{i,t-1}^{-1}\bb_{i,t-1}$, \quad $i = 1, \ldots, n$;
\STATE Get item set
       $
       C_{i_t} = \{\bx_{t,1},\ldots,\bx_{t,c_t}\};
       $
\STATE Determine $\hj_t \in \{1, \ldots, m_t\}$ such that $i_t \in {\hat U_{\hj_t,t}}$, and set
\vspace{-0.1in}
\begin{align*}
   \bM_{{\hj_t},t-1}  &= I+\sum_{i \in {\hat U_{\hj_t,t}}} (M_{i,t-1}-I),\\
   \bbb_{{\hj_t},t-1} &= \sum_{i \in {\hat U_{\hj_t,t}}} \bb_{i,t-1},\\
   \bbw_{{\hj_t},t-1} &= \bM_{{\hj_t},t-1}^{-1}\bbb_{{\hj_t},t-1}~;
\end{align*}
\vspace{-0.25in}
\STATE \[
       {\mbox{Set\ \ }}
        k_t = \argmax_{k = 1, \ldots, c_t} \left({\bbw_{\hj_t,t-1}}^\top\bx_{t,k}  + \CB_{\hj_t,t-1}(\bx_{t,k})\right),
       \]
\vspace{-0.15in}
       \begin{align*}
       {\mbox{where\ \ }}
       \CB_{{\hj_t},t-1}(\bx) &= \alpha\,\sqrt{\bx^\top \bM_{{\hj_t},t-1}^{-1} \bx\,\log(t+1)}\,.
       \end{align*}
\vspace{-0.1in}
\STATE Observe payoff $a_t \in [-1,1]$;
\STATE Let ${\bar \bx_t} = \bx_{t,k_t}$;
\STATE Update weights:
       \begin{itemize}
       \vspace{-0.1in}
       \item $M_{i_t,t} = M_{i_t,t-1} + {\bar \bx_{t}}{\bar \bx_{t}}^\top$,
       \vspace{-0.1in}
       \item $\bb_{i_t,t} = \bb_{i_t,t-1} + a_t {\bar \bx_t}$,
       \vspace{-0.1in}
       \item Set $M_{i,t} =  M_{i,t-1},\ \bb_{i,t} = \bb_{i,t-1}$ for all $i \neq i_t$~;
       \end{itemize}
\STATE Update clusters:
\begin{itemize}
\vspace{-0.06in}
\item Flip independent coin $X_t \in \{0,1\}$ with $\Pr(X_t = 1) = r$.
       \begin{itemize}
       \vspace{-0.12in}
       \item If $X_t = 0$ then delete from $E_t$ all $(i_t,\ell)$ such that
       \[
             ||\bw_{i_t,t-1} - \bw_{\ell,t-1}|| > \sCB_{i_t,t-1} + \sCB_{\ell,t-1}~,
       \]
       \vspace{-0.3in}
       \begin{align*}
        \sCB_{i,t-1} &= \alpha\,\sqrt{\frac{1+\log(1+T_{i,t-1})}{1+T_{i,t-1}}},\\
        T_{i,t-1}    &= |\{s \leq t-1\,:\, i_s = i \}|,\qquad i \in \scU;
       \end{align*}
       \vspace{-0.28in}
       \item If $X_t = 1$ and $t \leq T/10$ then pick at random index $j \in \{1, \ldots, m_t\}$, $j \neq \hj_t$, and split $\hat U_{j,t}$ into two subclusters $\hat U_{j,t,1}$ and $\hat U_{j,t,2}$ by means of a standard graph clustering procedure. Delete all edges between $\hat U_{j,t,1}$ and $\hat U_{j,t,2}$.
\end{itemize}
\vspace{-0.15in}
\item Let $E_{t+1}$ be the resulting set of edges, set $G_{t+1} = (\scU,E_{t+1})$,
        and compute associated clusters
        $\hat U_{1,t+1}, \hat U_{2,t+1}, \ldots, \hat U_{m_{t+1},t+1}$~.
\end{itemize}
\ENDFOR
\end{algorithmic}
\caption{\label{alg:gclub}
Pseudocode of the GCLUB algorithm.
}
\end{center}
\end{figure}
\section{Experiments}\label{s:exp}
In this section, we briefly describe the setting and the outcome of our experimental investigation.

\subsection{Datasets}
We tested our algorithm on four freely available real-world benchmark datasets against standard bandit baselines for multiple users.

\textbf{LastFM, Delicious, and Yahoo datasets.}
For the sake of fair comparison, we carefully followed and implemented the experimental setup as described in~\cite{icml14} on these datasets, we refer the reader to that paper for details. The Yahoo dataset is the one called `` Yahoo 18k" therein.

\textbf{MovieLens dataset.} This is the freely available\footnote
{
http://grouplens.org/datasets/movielens~.
}
benchmark dataset MovieLens 100k.
In this dataset, there are $100,000$ ratings from $n = 943$ users on $1682$ movies, where each user has rated at least $20$ movies. Each movie comes with a number of features, like id, movie title, release date, video release date, and genres. After some data cleaning,
we extracted numerical features through a standard tf-idf procedure.
We then applied PCA 
to the resulting feature vectors so as to retain at least $95\%$ of the original variance, giving rise to item vectors $\bx_{t,k}$ of dimension $d=19$. Finally, we normalized all features so as to have zero (empirical) mean and unit (empirical) variance. As for payoffs, we generated binary payoffs, by mapping any nonzero rating to payoff 1, and the zero rating to payoff 0. Moreover, for each timestamp in the dataset referring to user $i_t$, we generated random item sets $C_{i_t}$ of size $c_t = 25$ for all $t$ by putting into $C_{i_t}$ an item for which $i_t$ provided a payoff of 1, and then picking the remaining 24 vectors at random from the available movies up to that timestamp. Hence, each set $C_{i_t}$ is likely to contain only one (or very few) movie(s) with payoff 1, out of 25. The total number of rounds was $T = 100,000$. We took the first $10K$ rounds for parameter tuning, and the rest for testing.

\begin{figure}[t!]
\begin{picture}(22,25)(22,25)
\begin{tabular}{l@{\hspace{0pc}}l}
\includegraphics[width=0.52\textwidth]{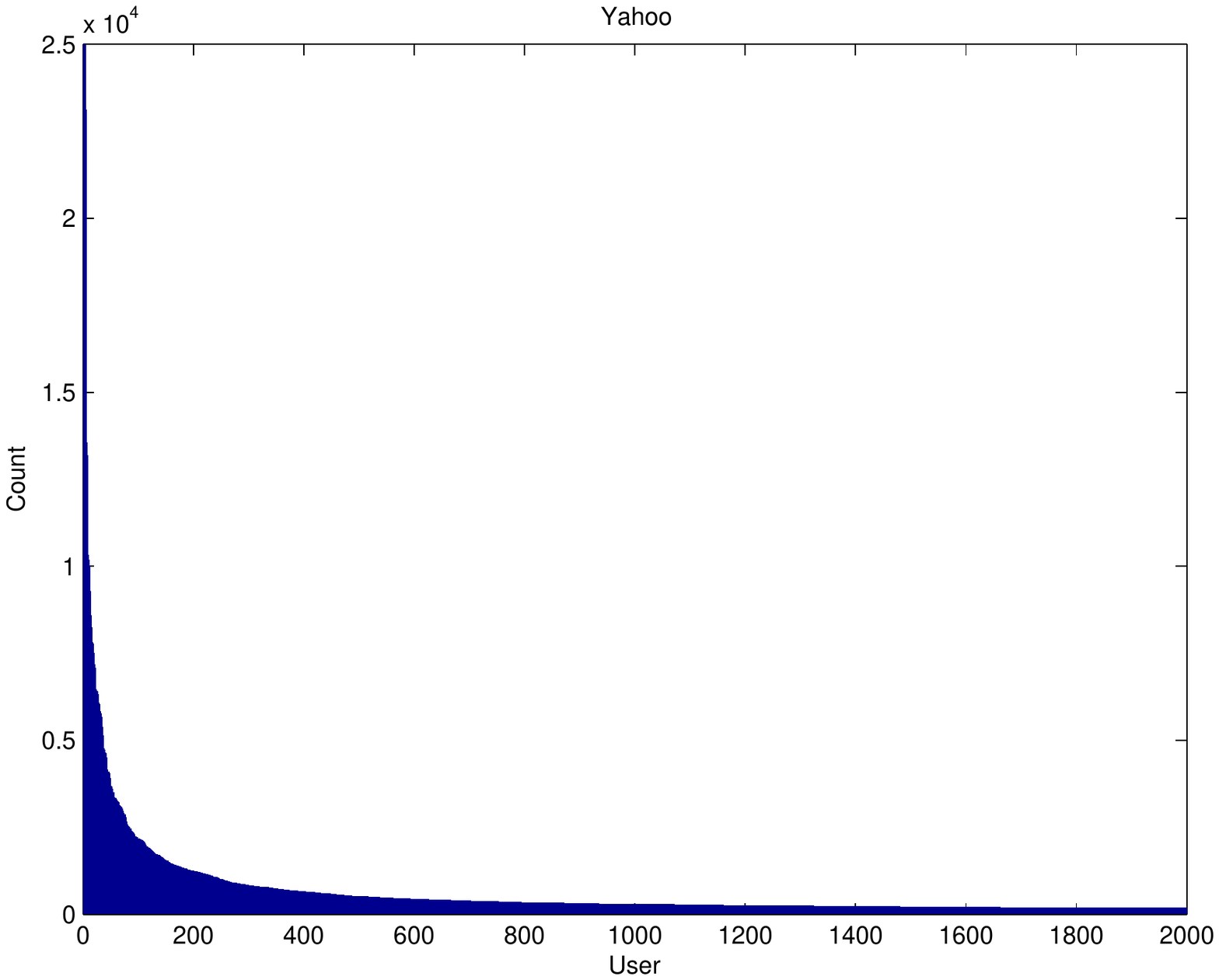}
& \includegraphics[width=0.52\textwidth]{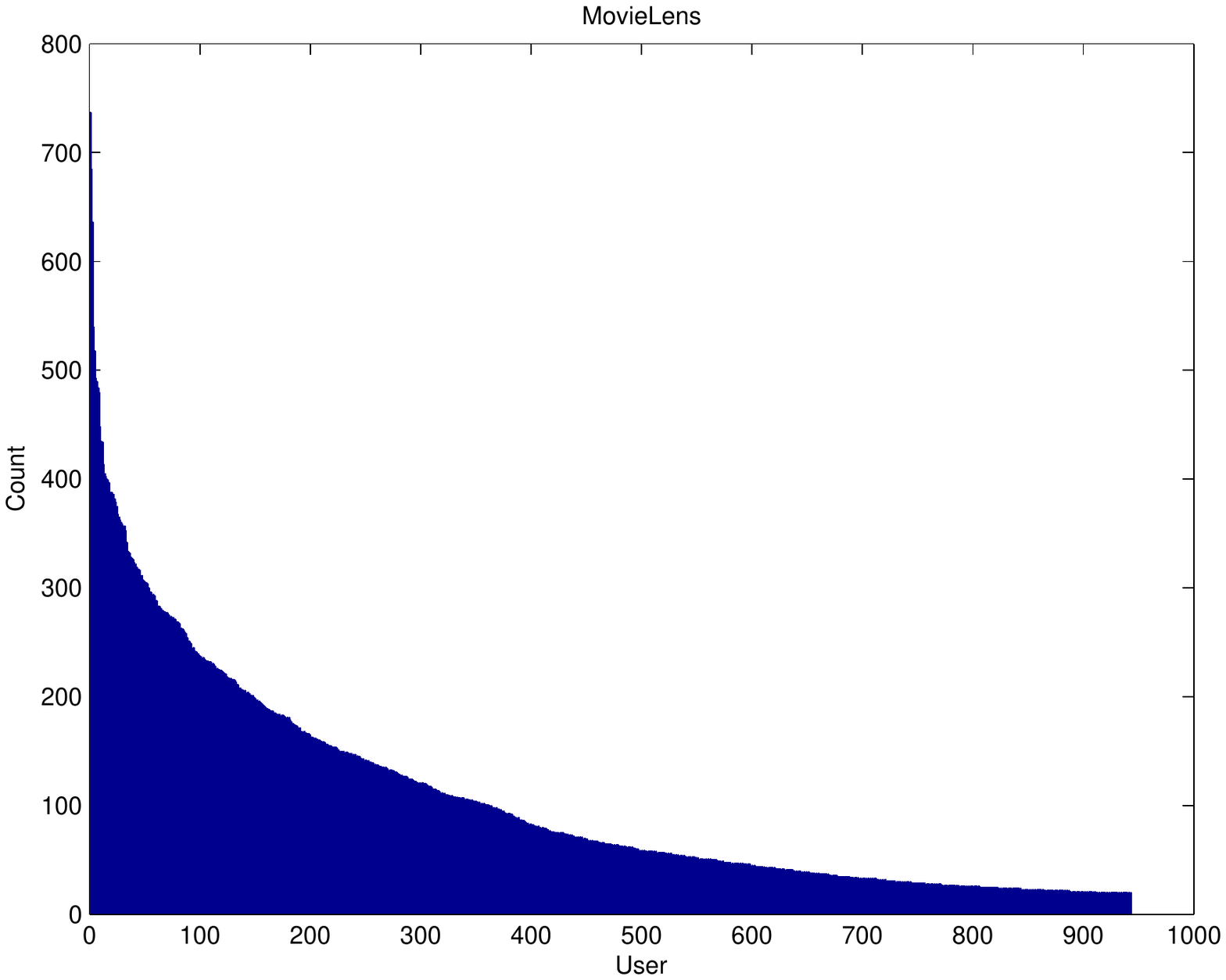}
\end{tabular}
\end{picture}
\vspace{1.4in}
\caption{\label{fig:activity_level} User activity levels on the Yahoo (left) and the MovieLens (right) datasets. Users are sorted in decreasing order according to the number of times they provide feedback to the learning system. For the sake of better visibility, on the Yahoo dataset we truncated to the 2K most active users (out of 18K).
}
\end{figure}

In Figure \ref{fig:activity_level}, we report the activity level of the users on the Yahoo and the MovieLens datasets. As evinced by these plots,\footnote
{
Without loss of generality, we take these two datasets to provide statistics, but similar shapes of the plots can be established for the other two datasets.
}
such levels are quite diverse among users, and the emerging pattern is the same across these datasets: there are few very engaged users and a long tail of (almost) unengaged ones.
%



\subsection{Algorithms}
We compared GCLUB to three representative competitors: LinUCB-ONE, LinUCB-IND, and CLUB.
LinUCB-ONE and LinUCB-IND are members of the LinUCB family of algorithms~\cite{Aue02,chu2011contextual,abbasi2011improved,cgz13} and are, in some sense, extreme solutions: LinUCB-ONE allocates a single instance of LinUCB across all users (thereby making the same prediction for all users -- which would be effective in a {\em few-hits} scenario), while LinUCB-IND (``LinUCB INDependent") allocates an independent instance of LinUCB to each user, so as to provide personalized recommendations (which is likely to be effective in the presence of {\em many niches}). CLUB is the online clustering technique from \cite{icml14}. On the Yahoo dataset, we run the featureless version of the LinUCB-like algorithm in~\cite{cgz13}, i.e., a version of the UCB1 algorithm of~\cite{ACF01}. The corresponding ONE and IND versions are denoted by UCB-ONE and UCB-IND, respectively. Finally, all algorithms have also been compared to the trivial baseline (denoted here as RAN) that selects the item within $C_{i_t}$ fully at RANdom.

We tuned the parameters of the algorithms in the training set with a standard grid search as in \cite{icml14}, and used the test set to evaluate predictive performance.
The training set was about 10\% of the test set for all datasets, but for Yahoo, where it turned out to be\footnote
{
Recall that on the Yahoo dataset records are discarded on the fly, so training and test set sizes are not under our full control --- see, e.g., \cite{lcls10,icml14}.
}
around 6.2\%. All experimental results reported here have been averaged over 5 runs (but in fact variance across these runs was fairly small).

\subsection{Results}
\begin{figure}[t!]
\begin{picture}(18,79)(18,79)
\begin{tabular}{l@{\hspace{0pc}}l}
\includegraphics[width=0.52\textwidth]{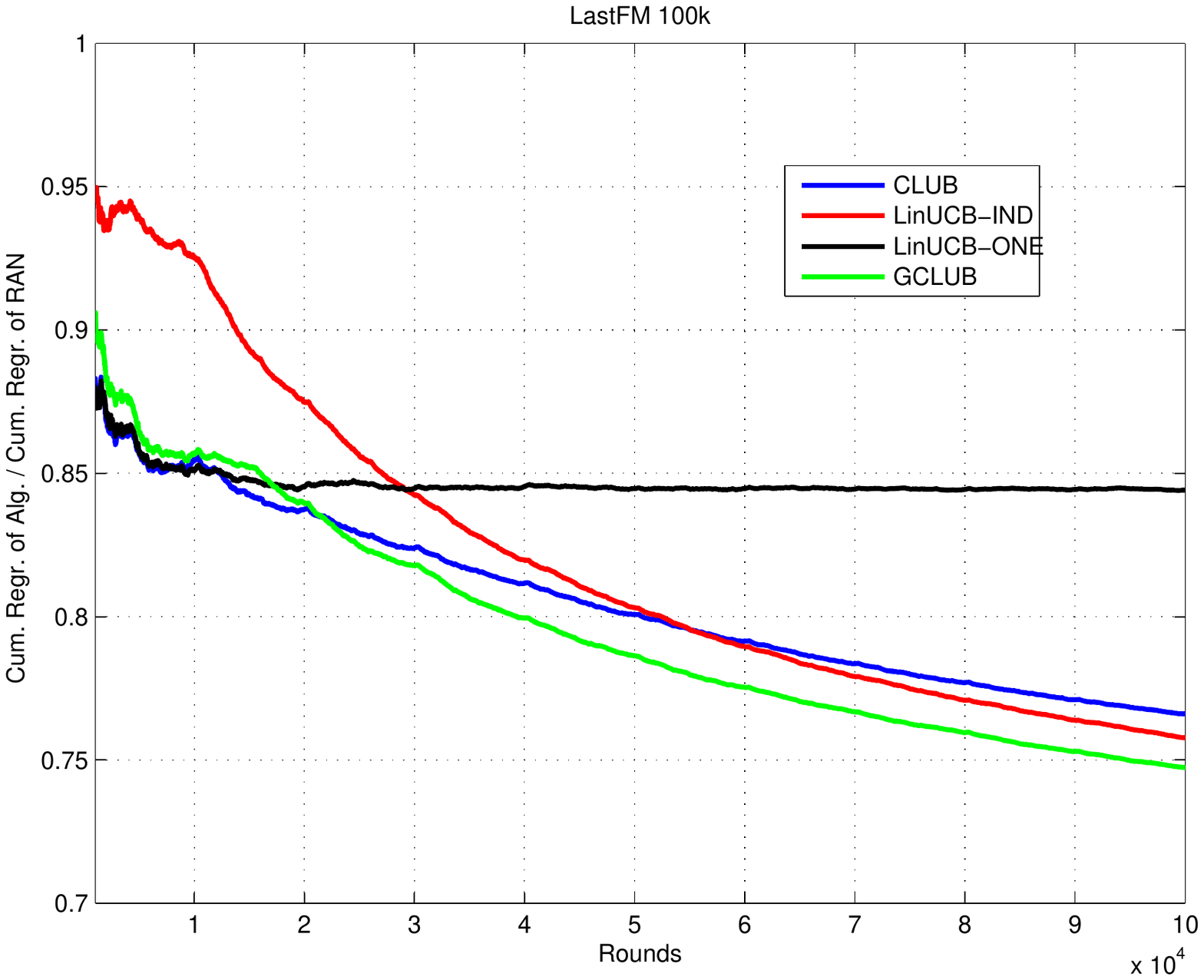}
& \includegraphics[width=0.52\textwidth]{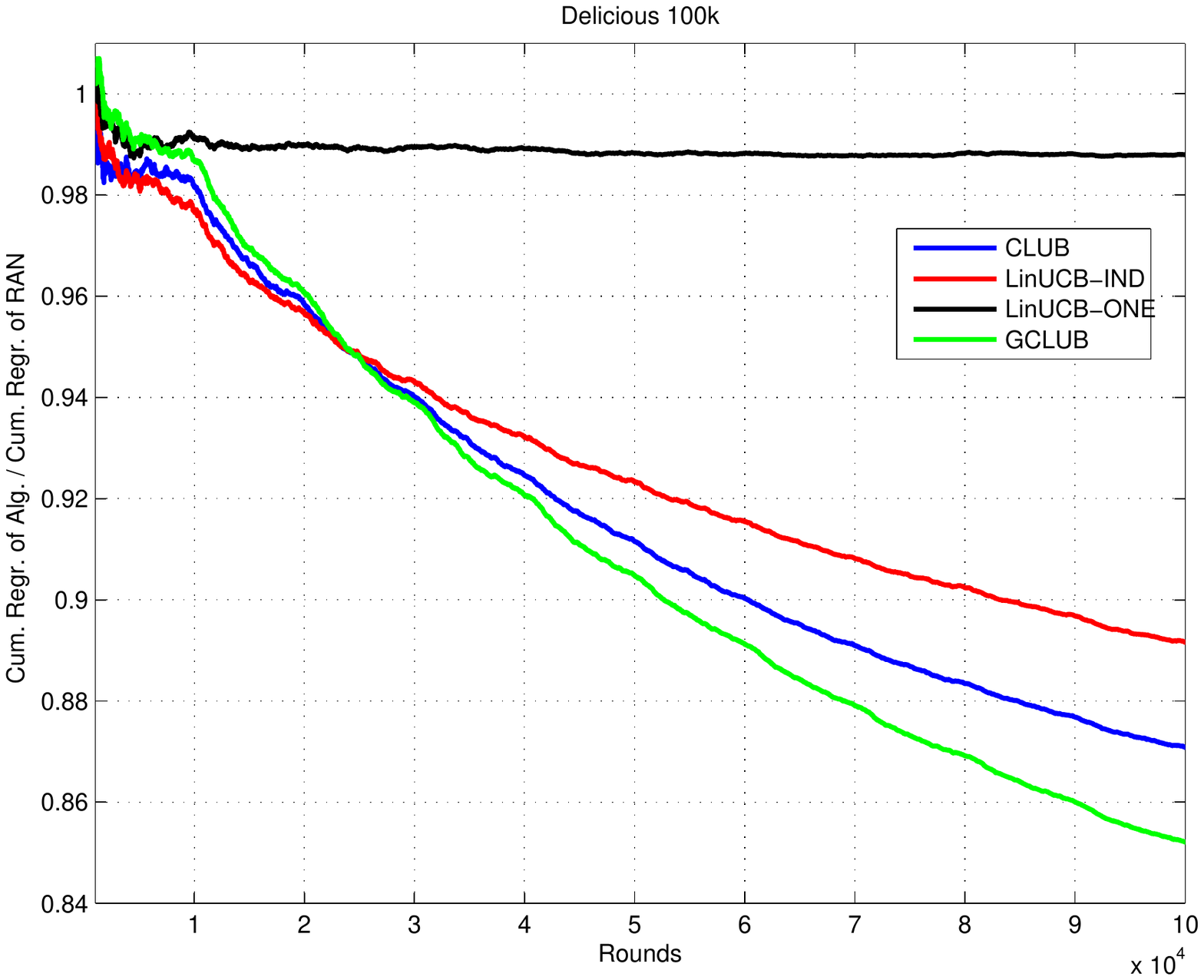}\\[-2.15in]
\includegraphics[width=0.52\textwidth]{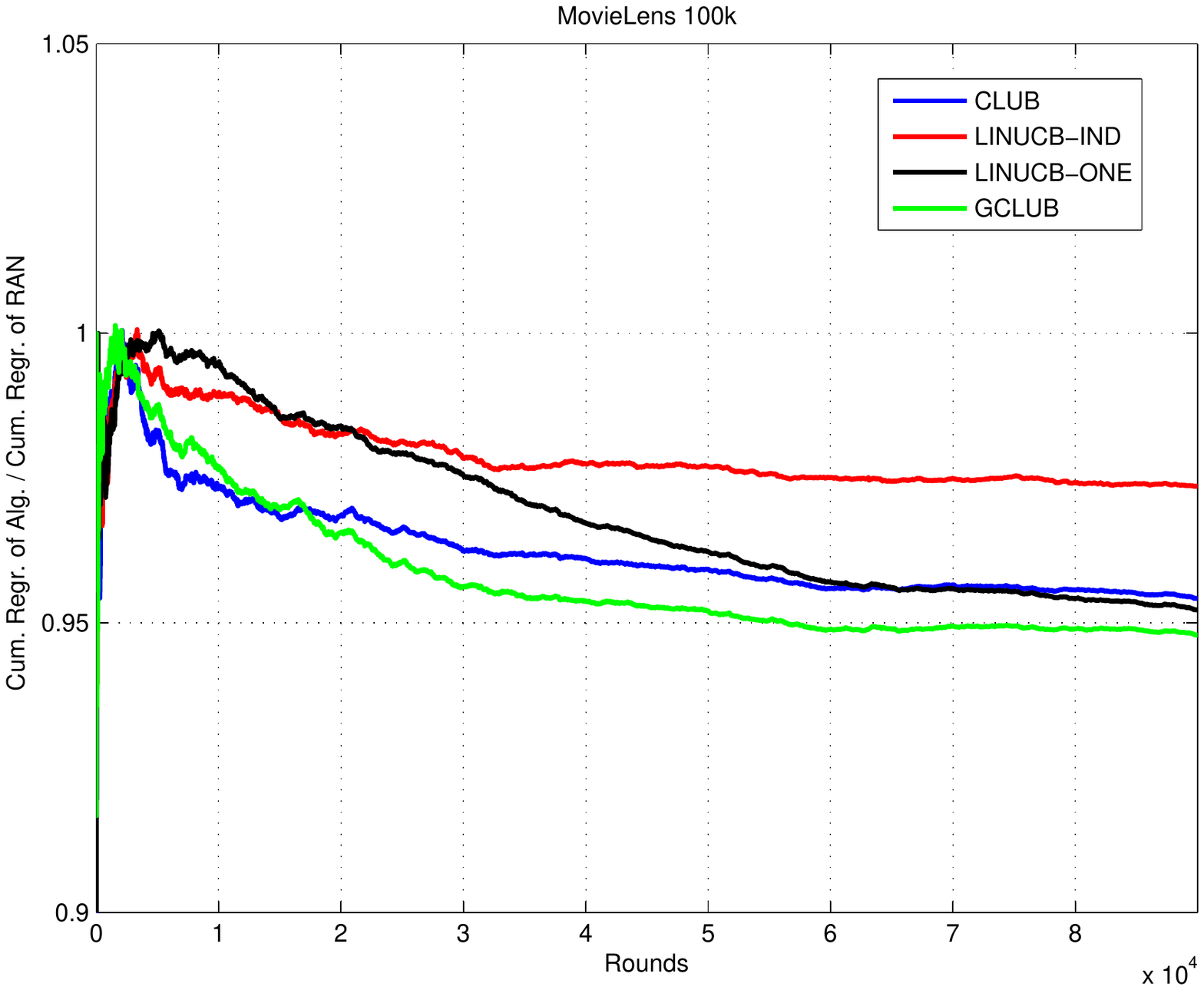}
& \includegraphics[width=0.52\textwidth]{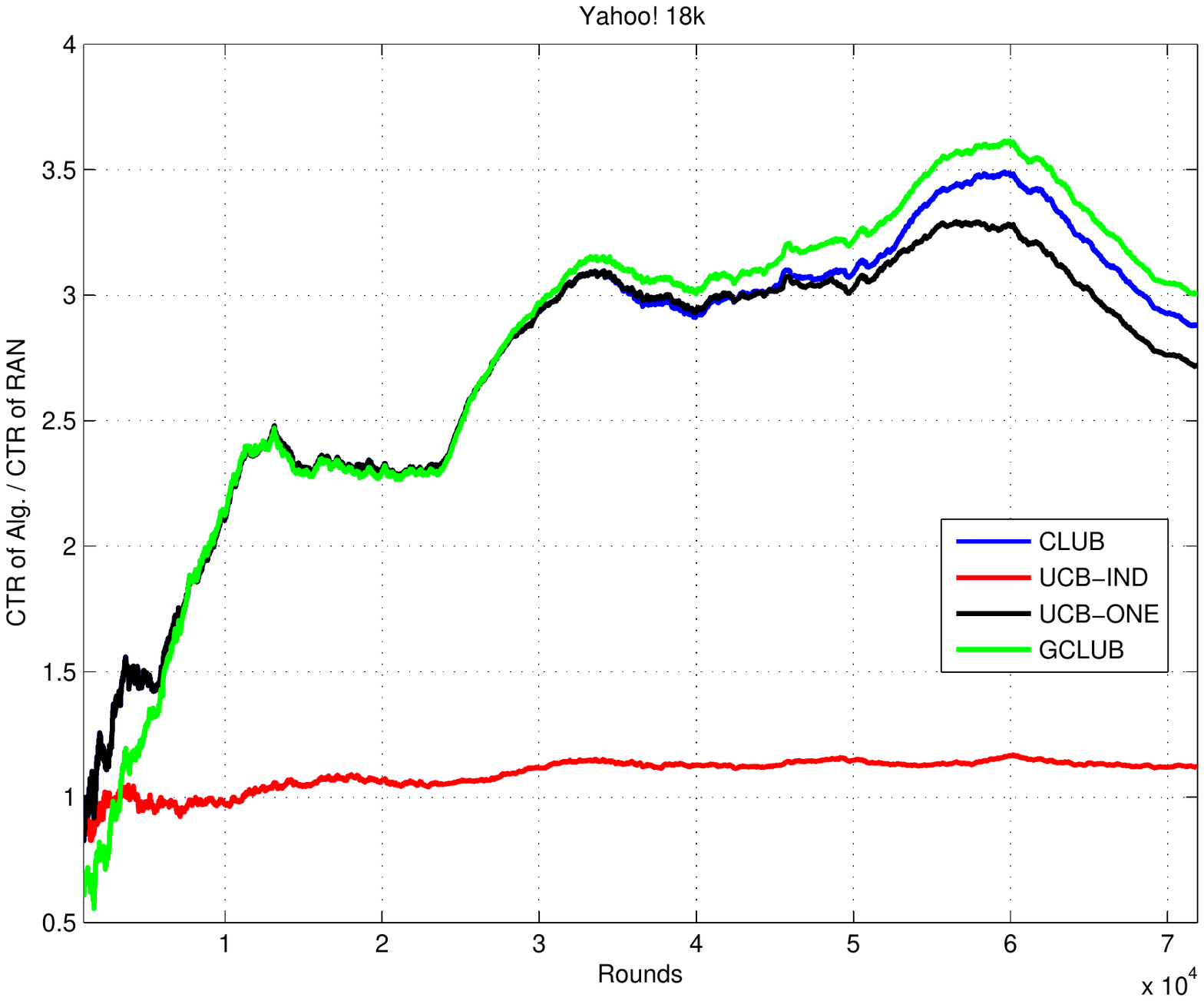}
\end{tabular}
\end{picture}
\vspace{3.1in}
\caption{\label{fig:last_del}Results on the LastFM (top left), the Delicious (top right), MovieLens (bottom left), and the Yahoo (bottom right) datasets. On the first three plots, we display the time evolution of the ratio of the cumulative regret of the algorithm (``Alg") to the cumulative regret of RAN, where ``Alg" is either ``GCLUB" (green), CLUB (blue), ``LinUCB-IND" (red), or ``LinUCB-ONE" (black). On the Yahoo dataset, we instead plot
the ratio of Clickthrough Rate (CTR) of ``Alg" to the Clickthrough Rate of RAN. Colors are consistent throughout the four plots.
}
\end{figure}

Our results are summarized in Figure \ref{fig:last_del}, where we report test set prediction performance. On LastFM, Delicious, and MovieLens, we measured the ratio of the cumulative regret of the algorithm to the cumulative regret of the random predictor RAN (so that the lower the better). On the Yahoo dataset, because the only available payoffs are those associated with the items recommended in the logs, we measured instead the ratio of Clickthrough Rate (CTR) of the algorithm to the CTR of RAN (so that the higher the better).

Whereas all four datasets are generated by real online web applications, it is worth remarking that these datasets are indeed quite different in the way customers consume the associated content. For instance, the Yahoo dataset is derived from the consumption of news that are often interesting for large portions of users, hence there is no strong polarization into subcommunities (a typical ``few hits" scenario). It is thus unsurprising that on Yahoo (Lin)UCB-ONE is already doing quite well. This also explains why (Lin)UCB-IND is so poor (almost as poor as RAN). At the other extreme lies Delicious, derived from a social bookmarking web service, which is a {\em many niches} scenario. Here LinUCB-ONE is clearly underperforming. On all these datasets, CLUB performs reasonably well (this is consistent with the findings in \cite{icml14}), but in some cases the improvement over the best performer between (Lin)UCB-ONE and (Lin)UCB-IND is incremental. On LastFM, CLUB is even outperformed by LinUCB-IND in the long run. Finally, GCLUB tends to outperform all its competitors (CLUB included) in all cases.

Though preliminary in nature, we believe these findings are suggestive of two phenomena: i. building clusters over users solely based on past user behavior can be beneficial; ii. in settings of highly diverse user engagement levels (Figure \ref{fig:activity_level}), combining sequential clustering with a stochastic exploration mechanism operating at the level of cluster formation may enhance prediction performance even further.


%
\bibliographystyle{abbrv}
\bibliography{biblio}  
%
%

\end{document}